\documentstyle[twocolumn,a4wide,lsalike,gb4e]{article}

\newcommand{\ul}{\underline}

\newcommand{\border}{\rule{\textwidth}{0.3mm}}

\title {\bf Toward an MT System without Pre-Editing\\
       --- Effects of New Methods in ALT-J/E ---}

\author { Satoru {\sc Ikehara}, Satoshi {\sc Shirai}, \\
          Akio {\sc Yokoo} and Hiromi {\sc Nakaiwa}  \\
          NTT Communications and Information Processing
          Laboratories\thanks{{\it Now:\/}  NTT Communication Science
            Laboratories.}\\
          1-2356 Take, Yokosuka-shi, Kanagawa-ken, {\sc Japan}  238-03 \\
          E-mail: ikehara@nttkb.ntt.jp}
\date{{\bf MT Summit III, 1991}}

\begin{document}

\maketitle

\begin{abstract}

  Recently, several types of Japanese-to-English machine translation
  systems have been developed, but all of them require an initial
  process of rewriting the original text into easily translatable
  Japanese.  Therefore these systems are unsuitable for translating
  information that needs to be speedily disseminated.  To overcome
  this limitation, a Multi-Level Translation Method based on the
  Constructive Process Theory has been proposed.  This paper describes
  the benefits of using this method in the Japanese-to-English machine
  translation system {\bf ALT-J/E}.

  In comparison with conventional compositional methods, the
  Multi-Level Translation Method emphasizes the importance of the
  meaning contained in expression structures as a whole.  It is shown
  to be capable of translating typical written Japanese based on the
  meaning of the text in its context, with  comparative ease.
  We are now hopeful of carrying out useful machine translation with
  no manual pre-editing. \end{abstract}

\section {Introduction} 

Recently, R\&D efforts involving machine translation of different
language families, such as Japanese and English, have become popular
\cite{Tomabechi:1987,Tomita:1987,Summit:1987}.  However, differences
in perspective and how objects are thought of, in such different
language families, affect how expressions are structured.  These
differences in expression structure make it difficult to convert from
one language to another mechanically.  For example, in
Japanese-to-English machine translation, the more typical the Japanese
expression, the more difficult it is to translate into English, due
to differences in the way that thought processes are expressed
linguistically.

As a means of solving this problem, efforts have been made in the area
of sub-languages \cite{Nagao:1985} or knowledge-based translation
\cite{Nirenburg:1989}.  But these methods currently require human
intervention, that is, Japanese expressions must be rewritten into
easily translatable Japanese.  In other words, there is a need to
re-write the text into a more English type of concept before machine
translation can be performed.

This action of re-writing is normally known as pre-editing
\cite{Nagao:1989}.  Pre-editing techniques include: use of a single
word so as to have only one meaning; limiting the use of `joshi'
(Japanese post-positional words) and of auxiliary verbs and other words
likely to be interpreted several ways; replacing, in advance, any
words which may have been omitted; and the re-writing of idiomatic
expressions to more general expressions.  These all represent efforts
to re-write the source into unambiguous Japanese which can be
translated into English, literally.

The theory and rationale of pre-editing in Japanese-to-English
translation would appear to be closely related to the principle of
elementary compositionality.  Elementary compositionality hypothesizes
that ``the meaning of the entire expression is the sum of the meanings
of the various portions of the expression'' \cite{Nomoto:1986}.  This
principle is taken as basic in existing machine translation systems
and supports a most effective method between languages of the same
family.  When seeking high quality machine translation, however, there
still remain serious problems to be dealt with.

Japanese-to-English machine translation has reached the stage where
sentences that allow word-by-word transfer from Japanese to English,
followed by assembly into the final sentence form (i.e. where literal
translation is possible), can be translated by current technology. But
there is a wide difference in the thought process constituting the
background of linguistic expression between the Japanese and English
languages.  Therefore, translations using existing systems require
pre-editing to re-write the original Japanese sentences into a form
that will enable application of the elementary compositional method,
or in other words, a form that can undergo literal translation.

To go beyond the limits of conventional translation methods based on
elementary compositionality, we have proposed the Multi-Level
Translation Method, \cite{Ikehara:1987,Ikehara:1989a,Ikehara:1989b}
based on the Constructive Process Theory of Language
\cite{Tokieda:1941}, and have made the experimental system {\bf
ALT-J/E}, the Automatic Language Translator --- Japanese to English.

This method focuses attention on the fact that many expressions have
meanings that cannot be deduced directly from the combination of the
meanings of the individual words.  It is a method of translation which
grasps the structure and meanings of expressions as a whole.  The
meanings of words will vary according to the manner and context in
which the words are used.  Many expressions have meanings that cannot
be explained directly from the meanings of each individual word.  With
attention focused on these characteristics, those units having
structural meanings have been arranged systematically into a form of
linguistic knowledge. This knowledge is being used in analysis of
the Japanese language and conversion of the Japanese into English.  As
such, it represents a big step forward towards the fundamental solution of
previously existing problems, hitherto only solvable by pre-editing.

\section{The Constructive Process Theory and the Multi-Level Translation
Method} 

\subsection  {The Constructive Process Theory of Language} 

\subsubsection{Problems of Conventional Translation Systems}

The transfer method and the pivot method have been regarded as the
methods most commonly used in machine translation \cite{Summit:1987}.
Whereas the pivot method hypothesizes an intermediate language common
for both the original and the target language, the transfer system
differs in that it uses an intermediate language for each language in
order to convert meanings from one language to another.  Both have in
common the fact that they establish an intermediate language to
represent meaning that is separate from the surface expression.

It is possible to seek the background regarding these methods in the
dualism of computational linguistics
\cite{Chomsky:1956,Chomsky:1965,Fillmore:1975} that discriminates
between surface and deep structures.

But the deep structure as suggested by computational linguistics
cannot be said to have achieved success.  In fact, concepts which deny
the existence of deep structure have been suggested of late
\cite{Cresswell:1973,Mendelson:1979,Bresnan:1982}.

Computational linguistics can be thought of as derived from
computational logic \cite{Allwood:1971}.  It hypothesizes that
the meanings of expressions do not rely on languages but are a form of
common existence, and it also hypothesizes that the meaning of the
expression in its entirety is the sum total of the meanings of
sections of the expression.  But these hypotheses are only partially
valid for actual languages.  Thus, it would be difficult to apply this
theory of computational linguistics to machine translation which deals
with actual text, particularly to translation involving a pair of
languages with different origins such as Japanese and English.

\subsubsection{The Concept of the Constructive Process Theory of Language} 

We believe that the key to solving this problem lies with the linguistic
evolution theory of Tokieda Grammar \cite{Tokieda:1941}, one of the
main schools of traditional study of the Japanese language.
Tokieda Grammar is derived from the theory of Norinaga Motoori
\cite{Motoori:1779} and it was developed from a critique of the
linguistic theory propounded by Saussure \cite{Saussure:1909}.  It is
regarded as one of the 4 major theories of grammar of Japan.

According to the Constructive Process Theory of Language, language is
to be understood as a compound body of processes as in the field of
physics, and can be viewed as the relationship between the `object',
`(speaker's) recognition' and `expression'.  The relationship between
`object' and `recognition' can be explained by `Epistemology' or
`Reflection Theory', and between `recognition' and `expression' by
`Linguistic Norm'.  The sole element that is common between two
differing languages would be the `object' and since there are
differences in how the `object' is viewed and understood between
languages, everything beyond `recognition' will differ depending on
the language in question.  The very existence of `deep structure'
which is neither `object' nor `recognition' is denied altogether.

Also, according to Tsutomu Miura \cite{Miura:1967} who built on the
Constructive Process Theory, the meaning of linguistic expressions is
the relationship between `object', `recognition' and `expression'.
This relationship is objectively connected to the `expression' itself.
The concept of regarding ``relationship" as meaning resembles the
recent work in situation semantics \cite{Barwise:1981}.  But where
situation semantics confuses ``meanings of expression'' with ``meanings
of the field where the expression is placed'', Miura Grammar draws a
distinct line between the two and propounds a theory pertaining to
``meanings of expression".

When language is regarded thus as a compound body of various
processes, the following two points, placing importance on the
meaning, are seen as important for machine translation.

\begin{enumerate}
\item Expressions are classified\footnote{ Regarding the difference
    between subjective and objective expressions, there is the theory
    of Port Royal \cite{Lancelot:1660}, before Norinaga Motoori.} into
  `subjective' which are a direct expression of the emotions,
  intentions, and judgment of the speaker and `objective' which
  express the object in the form of a concept, and reproduce it within
  the framework of the target language.

\item The structure, which involved with the object, is reflected by
  its recognition and this is further reflected in the structure of
  the expression.  Therefore, the structure of an expression is to be
  considered as a part of its meaning, and the meaning is to be handled
  accordingly.
\end{enumerate}

\begin {figure*}[t]
\border
\begin {center}
  \setlength{\unitlength}{1mm}
  \begin{picture}(140,80)
    \put(22,76.5){\oval(24,7)}
    \put(11,73){\makebox(22,6){\bf Japanese}}
    \put(22,73){\vector(0,-1){8}}
    \put(118,76.5){\oval(24,7)}
    \put(107,73){\makebox(22,6){\bf English}}
    \put(118,65){\vector(0,1){8}}
    \put(5,55){\framebox(35,10){\shortstack{Japanese\\Analysis}}}
    \put(40,60){\vector(1,0){10}}
    \put(22,55){\vector(0,-1){12}}
    \put(50,55){\framebox(40,10){\shortstack{Subjective Expression\\Transfer}}}
    \put(90,60){\vector(1,0){10}}
    \put(100,55){\framebox(35,10){\shortstack{English\\ Generation}}}
    \put(22,38){\oval(24,10)}
    \put(11,33){\makebox(22,10){\shortstack{Objective\\Expression}}}
    \put(34,38){\vector(1,0){9}}
    \put(118,38){\oval(24,10)}
    \put(107,33){\makebox(22,10){\shortstack{Objective\\Expression}}}
    \put(97,38){\vector(1,0){9}}
    \put(118,43){\vector(0,1){12}}
    \put(43,44){\makebox(54,8){\bf The Multi-Level Transfer Method}}
    \put(43,35){\framebox(54,6){\shortstack{Idiomatic Expression Transfer}}}
    \put(70,35){\vector(0,-1){5}}
    \put(43,24){\framebox(54,6){\shortstack{Semantic Valency Transfer}}}
    \put(97,25){\vector(1,0){5}}
    \put(70,24){\vector(0,-1){5}}
    \put(43,13){\framebox(54,6){\shortstack{General Pattern Transfer}}}
    \put(97,16){\line(1,0){5}}
    \put(102,16){\vector(0,1){22}}
    \put(35,70){\makebox(70,8){\shortstack{\bf Separation and
          Recombination\\\bf of Subjective and Objective parts}}}
    \put(2,52){\dashbox(136,16){}}
    \put(40,10){\dashbox(60,34){}}
  \end{picture}
  \vspace{-15mm}
\end {center}
\caption{The Multi-Level Translation Method}
\label{fig:outline}
\border
\end{figure*}
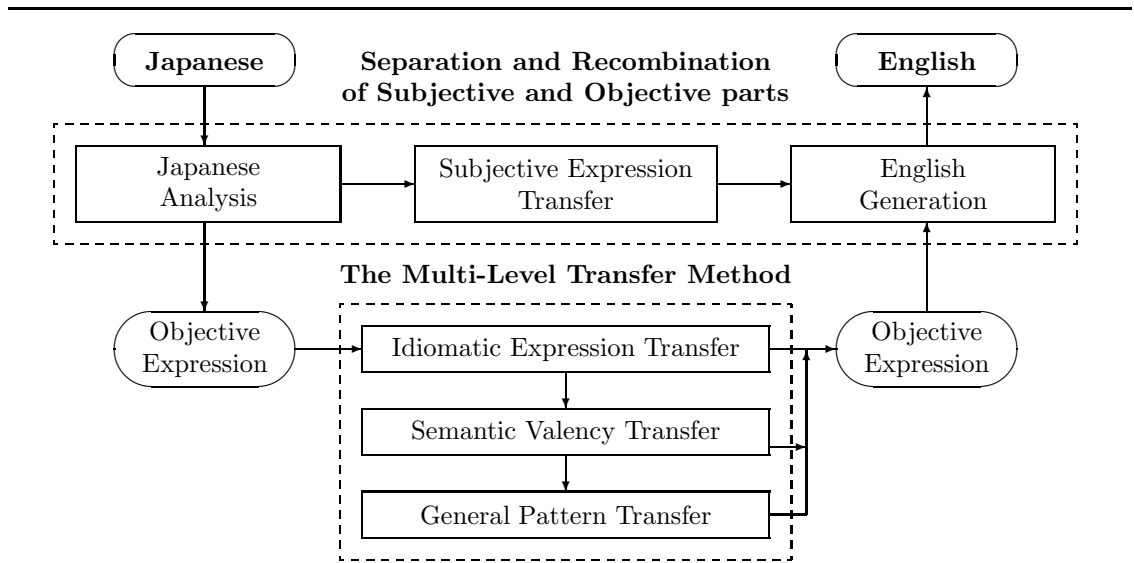

\subsection {The Multi-Level Translation Method} 

{\bf ALT-J/E} has implemented the Multi-Level Translation Method with
due consideration of the foregoing two points.  The translation
process is outlined in Figure~\ref{fig:outline}.  First, the Japanese
expression is analyzed and separated into subjective and objective
parts.  The subjective part (for example, tense and aspect) is
translated separately from the objective part.  Second, the objective
part is translated in three stages (the Multi-Level Transfer Method).
If there are any idiomatic expressions, these are translated first, in
the Idiomatic Expression Transfer.  Then any expressions whose
predicates and arguments match an entry in the semantic pattern
dictionary are translated as part of the Semantic Valency Transfer.
Finally, any remaining expressions are translated by the General Pattern
Transfer.  The entire process is designed to prevent loss of meaning
through elementary decomposition.

\section{Organization of Linguistic Knowledge} 

\subsection{Semantic Categories of Words} 

Nouns are used to express existing objects as concepts.  Depending on
how the object is viewed and understood, various profiles of the
object are picked up or discarded.  Which noun is to be used is
selected based on a profile corresponding to the view of the speaker.

In conceiving the object, the special and individual characteristics
are discarded and the features are recognized as a single unit.  Among
the concepts analyzing semantic features, there have been attempts to
explain the meaning of nouns as a bundle of detailed meanings or
features.  But the concept that is represented by a noun is a single
conclusive unit of recognition.  It is, therefore, to be handled as an
irreducible concept, that can only be captured as a whole.  We
classify these concepts as {\sc semantic categories}.

For example, the objective concept represented by the word {\it
  school\/} would include ``the school as an organization'' and ``the
school as a given location''. In machine translation, there is a need
to know which of these the word {\it school\/} signifies.  In order to
do this, thought was given to what type of profile is conceived for
the object when it is used. These profiles were then classified as
semantic categories held by each noun.

Around 3,000 categories were specified, about the number of important
words which the normal person feels comfortable in using.  The
semantic categories are ordered into two {\sc is-a} hierarchies. These
are the common noun semantic categories, some 2,800 categories (12
levels deep), and the proper noun semantic categories, some 200
categories (9 levels deep).  Based on this ontology, a {\bf semantic
  word dictionary} was compiled with 400,000 index words.  The maximum
number of semantic categories per word is 5 common noun categories and
10 proper noun categories.  Overall, an average of 2 categories are
assigned to each noun in the dictionary.

Projects using conceptual classifications similar to our semantic
categorization, have previously had around 30 to 50 categories. EDR
\cite{EDR:1990}, is implementing plans to extend to 500 categories. {\bf
ALT-J/E} is the first case of a system with a precision of some 3,000
categories and a large scale dictionary (around 400,000 index words).

\subsection  {The Meaning of Expression Structures
as viewed from Declinable Words} 

In Japanese both verbs and adjectives are declinable.  The basic
structure of Japanese sentences revolves mainly around
predicates.  Looking at the declinable words, the meanings of the
predicates themselves, and of their basic structure, can be understood
by examining the types and meaning of nouns that fill the predicate's
case frames. A {\bf semantic structure dictionary} with some 6,000
index words (verbs and adjectives) consisting of 15,000 patterns has
been prepared for use in analysis, transfer and generation.

With this method, analysis is performed by having units of semantics
and structure correspond to one another so that ambiguity in
structural analysis is reduced.  Each Japanese entry has an English
translation. As soon as the structure of the Japanese is determined in
the source language analysis, the basic English structure can be
determined from the English form structure in the semantic structure
dictionary.  This is helpful in avoiding the need for an additional
conversion process.

\section {Realization of New Functions} 

Among the functions which have been realized through this method, the
following  will solve problems previously requiring pre-editing.

\subsection {Precise Selection of Translation
  According to Meaning} 

Previously, re-writing of the original text was required so that each
word in the source would have an unambiguous translation in the
target language.  But, due to the rich information in the semantic
structure and word dictionaries, it has now become possible to
differentiate into precise translations as shown in Figure~\ref{fig:differ}.
Manual rewriting is no longer necessary.

\begin{figure*}[htbp]
\border

\small \begin{tabular}{ll}
\multicolumn{2}{c}{\normalsize \bf Differentiating translation of the verb
 {\it kakeru\/} `hang'} \\
{\it kanojo-wa hana-ni \ul{mizu-o kaketa}.} \dotfill &
She \ul{poured water} on a flower. \\
{\it haha-wa kamisama-ni \ul{gan-o kaketa}.} \dotfill &
A mother \ul{made a vow} to God. \\
{\it watashi-wa karera-ni \ul{meiwaku-o kaketa}.} \dotfill &
I \ul{caused} them \ul{trouble}. \\
{\it kare-wa nikai-ni \ul{hashigo-o kaketa}.} \dotfill &
He \ul{placed a ladder up to} the second floor. \\
{\it kensetsush\=o-wa koko-ni \ul{hashi-o kaketa}.} \dotfill &
The Ministry of Construction \ul{built a bridge} here. \\
{\it kare-wa isu-ni \ul{koshi-o kaketeiru}.} \dotfill &
He \ul{is sitting down} on a chair. \\
{\it karera-wa suna-o \ul{furui-ni kaketa}.} \dotfill &
They \ul{sifted} sand. \\
{\it kanojo-wa mainichi r\=oka-ni \ul{z\=okin-o kaketeiru}.} \dotfill &
She \ul{mops up} the corridor every day. \\
{\it kanojo-wa purezento-ni \ul{ribon-o kaketa}.} \dotfill &
She \ul{tied ribbon} around a gift. \\
{\it kanojo-wa shokutaku-ni \ul{t\=eburukurosu-o kaketa}.} \dotfill &
She \ul{spread a tablecloth} on a dining table. \\
{\it ano kissaten-wa \ul{modan-jazu-o kaketeiru}.} \dotfill &
That coffee shop \ul{is playing modern jazz}.
  \end{tabular}

  \begin{tabular}{ll}
\multicolumn{2}{c}{\normalsize \bf Differentiating translation of the noun
  {\it mure\/} `group'}\\
{\ul{\=okami-no mure-ga} \ul{hitsuji-no mure-o} otta.} \dotfill &
\ul{A pack of wolves} chased \ul{a flock of sheep}. \\
{\ul{kujira-no mure-ga} \ul{sakana-no mure-o} otta.} \dotfill &
\ul{A school of whales} chased \ul{a shoal of fish}. \\
{\ul{ushi-no mure-ga} \ul{hachi-no mure-ni} osowareta.} \dotfill &
\ul{A herd of cattle} was attacked by \ul{a swarm of bees}. \\
{\ul{hito-no mure-ga} \ul{b\=oto-no mure-ni} kawatta.} \dotfill &
\ul{A group of people} changed to \ul{a mod of a mob}.
  \end{tabular}

\caption{Precise Selection of Words in Translation}
\label{fig:differ}

\border
\end{figure*}

It has also become possible to translate typically Japanese
expressions which were previously difficult to translate into English
as well as to differentiate between translation of idiomatic
expressions and general expressions.

Further, it has become clear, after experimenting, that in order to
translate the meanings of Japanese declinable words (verbs and
adjectives) as shown in Figure~\ref{fig:differ} into English, it is
necessary to have a description of detailed rules. It has been
ascertained that this, in turn, requires a classification of detailed
semantic categories.  A look at rules involving the 15,000 cases
registered in the expression structure dictionary reveals the frequent
use of semantic categories classified in the 8th to 9th step in the
semantic category system.  This shows that at least the top nine
levels of our ontology (about 2,000 semantic categories) are needed to
successfully disambiguate most predicates.

\subsection  {Automatic Re-Writing Function in Japanese} 

There are many cases in which typical Japanese expressions, where two
or more words are combined to form idiomatic expressions, cannot be
literally translated and even if they were literally translated, would
be inappropriate in the English language.  It would be advantageous to
have such expressions automatically converted within the system into
more easily translatable Japanese.  But previous attempts to do this
have foundered due to the problems of unwanted side effects.

The Multi-Level Translation Method has enabled a precise enumeration
of conditions for the application of rules through detailed semantic
categories.  This has enabled side effects to be reduced and
effectively re-writes the Japanese prior to translation.

\begin{figure*}[htbp]

\border
\begin{exe}
\large \exi {\bf Original:}
\gll kare-wa basu-ni notte gakk\=o-e itta ga, watashi-wa kawa-ni
        sotte aruite gakk\=o-e itta.\\
        He bus ride school went but, I river {go along} walking school went.\\
\large \trans`He \ul{rode a bus} and went to school,
          but I \ul{paralleled the river},  \ul{walked} and
          went to school.' \\
\exi{\bf Rewrite:}
\gll kare-wa basu-de gakk\=o-e itta ga, watashi-wa kawa-zoi-ni
        toho-de gakk\=o-e itta.\\
        He {by bus} school went but, I {river along} {on foot} school went.\\
\trans   `He went to school \ul{by bus}, but I went to
        school \ul{on foot} \ul{along the river}.'

\end{exe}
\caption{Automatic Re-Writing in Japanese}\label{fig:rewrite}
\border
\end{figure*}

%
%


Figure~\ref{fig:rewrite} shows an example of a Japanese sentence, which
normally has numerous predicates but which has been automatically
rewritten so as to have fewer.  Three Japanese verb phrases are
changed into prepositional phrases in English.

\subsection{Supplementation of ellipsed elements through Context Processing}

The Japanese language normally omits elements that are easily
recoverable from context, particularly subjects and objects.  But in
English, these elements are in most cases obligatory.  Previously,
supplementing these constituted an important part of pre-editing.

ALT-J/E has, in addition to the semantic structure dictionary and
semantic categories, introduced an analysis of the semantic categories
of predicates which allows the supplemention of ellipses using the
semantic relations between sentences.

\subsection {Translation  of Compound Words} 

The Japanese language generates new words (compound words) which are
an amalgamation of a number of nouns, prefixes and suffixes (a
characteristic of agglutinative languages).  This type of compound
word is generated without limitation and so it is impossible to have
them all registered in a dictionary in advance.  With conventional
translation methods, registration of these compound words in the
dictionary was an important issue for pre-processing.

{\bf ALT-J/E} uses semantic categories to analyze compound words to
find the semantic relationships of their constituents.  This function
makes the translation of unknown compound words possible.  It also
enables the automatic translation of compound words whose meanings
vary  depending on the manner in which they are used within a
sentence.

\section {The Benefits of the Multi-Level Translation
Method and Future Issues} 

\subsection {Benefits of the Multi-Level Translation Method} 

The experimental Japanese-to-English machine translation system {\bf
  ALT-J/E}, based on the Multi-Level Translation Method, is currently
being debugged.  To examine the potential of this method, newspaper
lead sentences (a summary preceding the newspaper article proper,
generally consisting of 3 to 5 sentences per article, and averaging 20
words per sentence) were translated in the following experiments.

\begin {description}
\item [Blind Test:] (BT)\\ Experiments conducted with articles chosen
  at random with no registration of unknown words, nor rule revisions.
\end {description}

\begin {description}
\item [Window Test:] (WT)\\ Experiments on a sample of text with
  revision of the system allowed. Registration of unknown words and rule
  revisions are conducted during the test.
\end {description}

(In both cases, the original text was translated without any pre-editing) \\

The standards used for the evaluation are an improved version of the
ALPAC standards \cite{ALPAC:1966} with 10 points being a `perfect'
translation and grades 6 or higher being a pass (the sentence is
understandable from reading the translation only).  Grading was
conducted by outside company specialists in translation.  The average
of grades as judged by three specialists in Japanese-to-English
translation were taken to determine passing or failing grades for each
individual sentence.

The condition for a passing grade was that the meaning could be
understood by looking only at the translation.  Thus, sentences that
were ruled as passing are not guaranteed to be stylistically
appropriate (or even grammatical).  But it is estimated that a quality
level equal to or better than that of existing Japanese-to-English
machine translation systems has been achieved.

According to this test, the pass rates for the blind test were 40 to
50\% , and over 60\% for the window test. This indicates a passing
ratio of about double that of existing Japanese-to-English machine
translation systems.  For tests pertaining to technical subjects
(which are easier to translate than the newspaper lead sentences), a
pass rate of 80\% was achieved.

Based on the above results, we judge that with the Multi-Level
Translation Method, a major step toward realization of a
Japanese-to-English machine translation system requiring no
pre-editing has been achieved.

\subsection  {Future Issues} 

The major problem currently being faced is the need for improvement of
the translation quality of long sentences (of 30 words or longer) and
for the overall improvement of the English in the translated text.  To
meet this challenge, research efforts are presently being focussed on
an extended Japanese-to-English transfer method designed to analyze
the meaning of the structure of declinable words and to directly
establish an appropriate English structure to correspond to this.
This direct parse-tree transfer method will be adding a new path to
the three transfer paths for objective expression in the Multi-Level
Transfer Method, further improving and strengthening it.

Over the long term, research efforts are being extended to include a
review of the system of parts of speech in the Japanese language and
to extend the semantic hierarchy to multiple dimensions.

\section {Summary} 

This paper has presented the results of using the Multi-Level
Translation Method, based on the Constructive Process Theory. It has
shown that the method enables a Japanese-to-English machine
translation system to function effectively without manual pre-editing.
In fact, the major reasons for pre-editing the source text are no
longer valid. But there remain problems with translating typically
long Japanese sentences and a need to improve the quality of finished
translations.

We call the limited use of semantic information used in the
Multi-Level Translation Method {\bf meaning analysis}.  It is
estimated that this level of technology is limited to a maximum
success rate of approximately 80\%.  To attain a higher level of
accuracy it is essential to establish an understanding of meaning in
context, based on the expansion of general and specialized knowledge
of the target domains.  We call this {\bf meaning comprehension}.
However, since it is difficult to establish the comprehension of
meaning in extremely broad or general fields, it is planned to
establish the limits for processing based on meaning analysis first,
and then follow up with research into the area of meaning
comprehension.

\section* {Acknowledgment}

The authors wish to thank Masahiro Miyazaki, Kentaro Ogura
and other members of the research group on machine translation for
their valuable contribution to discussions; and especial thanks to
Francis Bond for revising this paper before archiving it.


\end{document}